\documentclass[runningheads]{llncs}
\usepackage[T1]{fontenc}
\usepackage{graphicx}
\graphicspath{{figures/}}
\usepackage{authblk}
\usepackage{amsmath}
\usepackage{amsfonts}
\usepackage{graphicx}
\usepackage[colorlinks=true, allcolors=blue]{hyperref}
\usepackage{bm}
\usepackage[ruled,vlined]{algorithm2e}
\usepackage{comment}
\usepackage{multicol}
\usepackage{paralist}
\usepackage{color}
\usepackage{subcaption}
\usepackage{fontawesome5}
\usepackage[utf8]{inputenc}
\usepackage{algpseudocode}
\usepackage{geometry}
\usepackage{tabularx}
\usepackage{float}
\usepackage{amsmath}
\usepackage{tabularx}  
\usepackage{threeparttable} 
\usepackage{soul}
\usepackage{booktabs} 
\definecolor{dkgreen}{rgb}{0,0.6,0}

\begin{document}
\newcommand{\R}{\mathbb{R}}
\newcommand{\todo}[1]{\textcolor{red}{+ (TODO) #1 +}}
\newcommand{\del}[1]{\textcolor{yellow}{- (del) #1 -}}
\newcommand{\new}[1]{\textcolor{blue}{#1}}  
\newcommand{\container}[1]{\textit{C}$_{#1}$}
\newcommand{\pu}[1]{\textit{P}$_{#1}$}
\newcommand{\timeincrement}{\delta}
\newcommand{\rmin}{r_{\min}}
\newcommand{\rpen}{r_\text{pen}}
\newcommand{\benchmark}{ContainerGym}

\title{Solving a Real-World Optimization Problem Using Proximal Policy Optimization with Curriculum Learning and Reward Engineering}

\titlerunning{Tackling Real-World Reinforcement Learning Deployment}

%
%
%
\author{Abhijeet Pendyala \faIcon{envelope} \and
Asma Atamna \and
Tobias Glasmachers
}
\authorrunning{A. Pendyala et al.}
%
%
\institute{Ruhr-University Bochum, Bochum, Germany\\
\email{firstname.lastname@ini.rub.de}
}

\maketitle  
\begin{abstract}

We present a proximal policy optimization (PPO) agent trained through curriculum learning (CL) principles and meticulous reward engineering to optimize a real-world high-throughput waste sorting facility. Our work addresses the challenge of effectively balancing the competing objectives of operational safety, volume optimization, and minimizing resource usage. A vanilla agent trained from scratch on these multiple criteria fails to solve the problem due to its inherent complexities. This problem is particularly difficult due to the environment's extremely delayed rewards with long time horizons and class (or action) imbalance, with important actions being infrequent in the optimal policy. This forces the agent to anticipate long-term action consequences and prioritize rare but rewarding behaviours, creating a non-trivial reinforcement learning task. Our five-stage CL approach tackles these challenges by gradually increasing the complexity of the environmental dynamics during policy transfer while simultaneously refining the reward mechanism. This iterative and adaptable process enables the agent to learn a desired optimal policy. Results demonstrate that our approach significantly improves inference-time safety, achieving near-zero safety violations in addition to enhancing waste sorting plant efficiency.

\end{abstract} 
\section{Introduction}{\label{intro}}

This work introduces a novel reinforcement learning approach to address the critical need for optimization within waste sorting facilities, addressing an uncharted area of research. EU directives emphasize the need for responsible and sustainable recycling of packaging waste. This has led to waste sorting facilities investing in sophisticated infrastructure and
automated sorting technologies. Traditional data-driven methods are emerging as key tools within the digitalization of these high-throughput facilities for optimal control. Sorting facilities are designed specifically for different use cases, such as the types of materials to be separated and the scale and the throughput requirements of the plant. They must remain robust to ever-changing material streams, and they should ideally be capable of adapting to changes of the input composition, and even to future design modifications such as changes in layout or modifying sorting machinery. Additionally, ensuring operational safety while optimizing sorting efficiency remains a key challenge.

The Reinforcement Learning (RL) framework allows agents to learn and optimize actions through dynamic interactions with their environments~\cite{sutton2018reinforcement}.  This makes it a powerful paradigm for creating optimal and adaptive control systems, particularly when dealing with uncertain or changing conditions \cite{Kaelbling1996-sh,Szepesvri2010}.  RL's ability to adapt based on feedback and reward signals offers great potential for addressing the complexities and ever-evolving nature of waste management processes.
The final stage of a waste sorting facility, namely the management of containers filling up with sorted material before being processed into a final product, is readily available as a real-world industrial benchmark RL environment called ContainerGym~\cite{Pendyala2024-vl}.

This research introduces a curriculum learning approach for training a Proximal Policy Optimization (PPO) algorithm for the container management problem. The approach breaks down the complex learning process into manageable stages, aiming to optimize sorting efficiency and resource utilization within operational constraints. 

The motivation behind our approach was that the RL problem at hand is non-trivial. We found that a vanilla baseline like a PPO agent trained from scratch on all the three criteria detailed in section~\ref{sec:Problem Description} fails to learn a useful control policy. Analysis of training data revealed that the majority of the training episodes terminated prematurely due to violation of safety constraints (volume limit exceeded). Consequently, the agent cannot collect sufficient samples of optimal state-action pairs, trapping them in a sub-optimal solution.
Therefore there was a need for a more powerful method. We demonstrate that a nuanced curriculum approach can solve the problem.

This paper is structured as follows. Section~\ref{sec:related_work} reviews relevant research and establishes the motivation for our approach. Section~\ref{sec:Problem Description} provides a detailed problem description and outlines key challenges. The control task is cast as a reinforcement learning (RL) problem in Section~\ref{sec:RL formulation}. Section~\ref{sec:reward_tuning} explains the design of a nuanced reward mechanism, on top of which Section~\ref{sec:methodolody} presents our structured curriculum learning strategy, composed of five distinct phases. Finally, Section~\ref{sec:results} presents an experimental evaluation and discusses results, followed by conclusions in Section~\ref{sec:conclusion}.

\section{Related Work}\label{sec:related_work}

Curriculum learning in the context of reinforcement learning is a training paradigm where an agent encounters a sequence of tasks carefully designed to accelerate learning. Designing a curriculum allows ML engineers and domain experts to incorporate experience and domain knowledge into the training process. The curriculum might involve modifying reward functions,  altering environment dynamics, or changing state or action spaces \cite{Svetlik2017-id}. By strategically guiding the agent's experiences, CL has the potential to promote faster convergence, better generalization, and more efficient learning \cite{Bengio_undated-ep,Narvekar2020-fn,weinshall2018curriculum}. Beyond reinforcement learning, CL has been proven to provide a versatile framework with a broad range of applications. In supervised learning, CL has enhanced model performance through the strategic sequencing of training examples \cite{Kumar_undated-ep,weinshall2018curriculum}. The gaming industry has employed CL to create progressively challenging levels, boosting agent performance \cite{Sukhbaatar2018-wx,Justesen2018-bh}. Robotics has benefited from CL, as complex tasks could be decomposed into simpler steps, accelerating robotic skill acquisition \cite{Florensa2017-bf,Riedmiller2018-rw}. Finally, CL has been used in safe reinforcement learning by gradually exposing the agent to higher-risk scenarios \cite{Garcia2012-rr,Garcia2015-kl}. For a comprehensive overview of CL methods, see the survey papers \cite{Wang2022-dc,Soviany2022-or,Narvekar2020-fn}.

Our proposed curriculum approach falls within the category of predefined curriculum learning \cite{Wang2022-dc}. We adopt a Curriculum Through Intermediate Goals strategy \cite{Gupta2021-pi}, decomposing a complex goal state into a sequence of simpler intermediate goals. While many existing approaches concentrate on modifying single aspects of the learning process, such as state distributions \cite{Florensa2017-bf}, reward functions \cite{Riedmiller2018-rw,wu2017training}, or goal generation \cite{Florensa2018,Racaniere2019}, we differentiate our method by strategically modifying multiple facets. We manipulate the underlying Markov Decision Processes (MDPs) of our intermediate tasks, carefully tuning environment dynamics, reward mechanisms, and learning time horizons. This multifaceted approach, combined with our focus on a real-world problem, sets our approach apart from existing methods. 

We aim to achieve three key benefits through our curriculum approach: optimization, 
improved sampling (as alluded to in \cite{Xu2022-st}), and safety. Optimization benefit arises as a curriculum guides an agent towards solutions more efficiently than direct minimization of a non-convex target objective with multiple criteria. Statistically, careful sample allocation within a curriculum can boost performance by facilitating knowledge transfer among tasks, potentially reducing training data needs. For safety, as \cite{Pollack2020-uy} note, a curriculum allows the reuse of safe policies from simpler tasks, ensuring safety in intermediate stages while the agent progressively develops proficiency for complex environments.

\section{Real Environment and Problem Description}\label{sec:Problem Description}

\begin{figure}
    \begin{center}
        \includegraphics[clip, trim=0.1cm 0.3cm 0.1cm 0.1cm,width=0.99\textwidth]{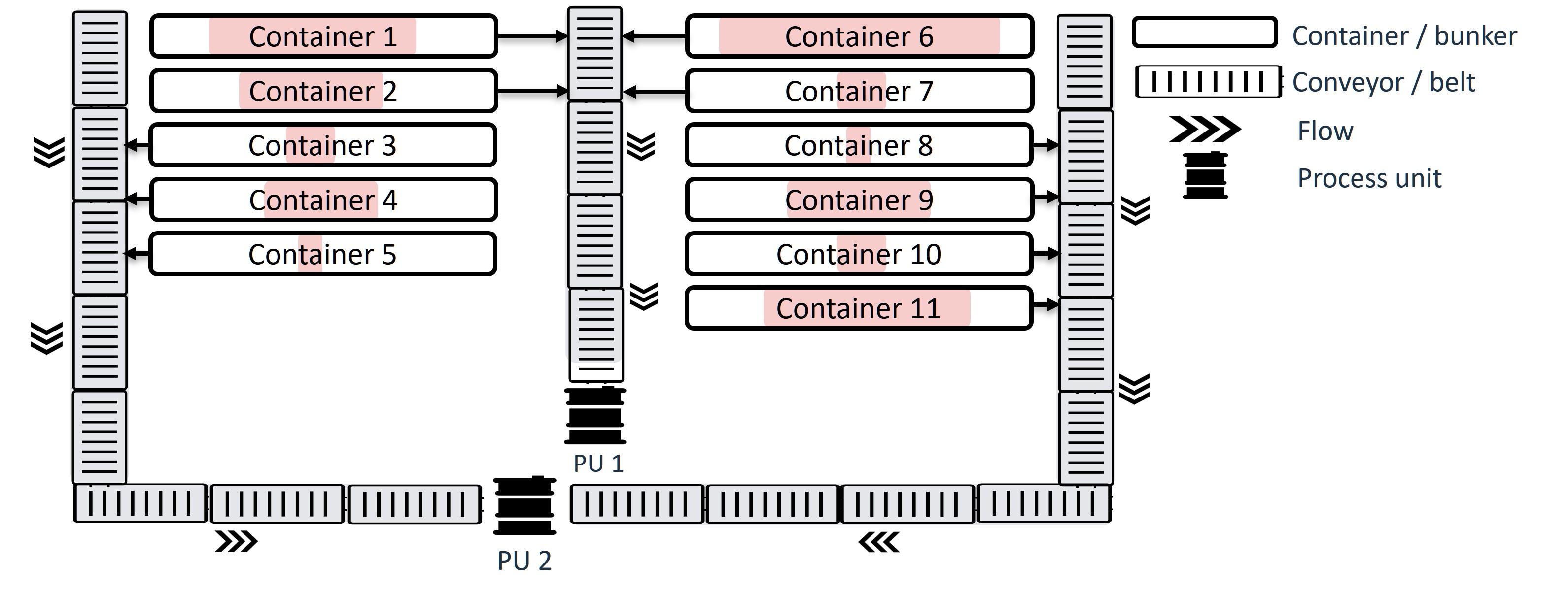}
    \end{center}
    \caption{\label{figure:task}
        Layout sketch of a facility with 11 containers and 2 PUs, connected with conveyor belts. The containers are filled from above, with their current fill states indicated by the shaded areas.
    }
\end{figure}

We consider the task of managing \emph{containers} in a plastic sorting facility. The containers collect pre-sorted material. After accumulating enough material of a certain type, the container is emptied onto a conveyor belt and the material is moved to one of two \textit{processing units} (PU-1 and PU-2). The setup is illustrated in Figure~\ref{figure:task}. Containers are continuously filled with material, with each container prescribed to a unique material. The material extraction from the container is always done until it is completely emptied. The difficulty of the task arises from two constraints. First of all, each container has its own unique optimal emptying volume depending on the material type. Emptying the container earlier is possible, but it results in a sub-optimal product and a waste of energy. Emptying later is subject to similar costs, however, processing larger chunks of material is generally beneficial as it saves energy and processing time. The second constraint is that processing the material takes time, and containers can be emptied only if a PU is free.

If a container is emptied too far away from its optimal (at higher and or lower value), the deviation in volume that cannot be transformed into a product is again redirected to the corresponding container via an energy-intensive auxiliary process. On the other hand, if a container is emptied prematurely and too frequently, the PUs are engaged too frequently causing higher energy usage and also a bottleneck in resource allocation. In other words, the control task is to reduce the cumulative deviation from the ideal volume for each container while managing resources (PUs) efficiently. 

A critical safety constraint that needs to be respected is that a container is never allowed to overshoot the physical limit of its maximum bearing volume (40 units). This incurs a high recovery cost including human intervention to stop and restart the facility. and should be avoided at all costs. Therefore, emptying containers close to the physical limit is a risky approach. Other system constraints are that certain containers can only be emptied into PU-1 and others only into PU-2. PU-1 takes in material via only one conveyor belt while PU-2 takes in material via two conveyor belts.

\noindent The quality of an emptying decision is a balance between three criteria:
\begin{compactitem}
    \item \textbf{Volume criteria}: minimize the cumulative deviation from the ideal emptying volume for each container.
    \item \textbf{Energy criteria}: optimize the energy usage or utilization costs of PUs.
    \item \textbf{Safety criteria}: adhere to the safety and system constraints. 
\end{compactitem}

\noindent Multiple aspects make this problem challenging:
\begin{compactitem}
    \item \textbf{Stochastic material flow rate.} The material flow rate into the containers is unique for each container, as it depends on the type of material, its density,  the time of the day, seasonality, and other factors. In addition, the sensor readings determining volume estimates are very noisy. Incorporating this phenomenon, the flow rates are modelled as a stochastic process. This makes approaching the problem with standard planning approaches quite difficult.
    \item \textbf{Delayed rewards.} Certain containers have very slow filling rates, taking up to 5 hours or 300 simulation timesteps of 60 seconds each to reach their respective ideal volumes, and receive a positive reward associated with the correct emptying action, as well as about 300 preceding `idle' actions.
    \item \textbf{PU constraint.} The limited availability of the PUs implies that always waiting for containers to fill up to their ideal emptying volume is risky: if too many containers are close to their respective ideal volumes and no PU is available at that time, then a safety constraint is violated and a physical overflow occurs. Therefore, an optimal policy needs to take fill states and fill rates of all containers into account, and possibly empty some containers early or later.
    \item \textbf{Class imbalance.} Emptying decisions can be taken at any time, but the emptying actions close to the ideal volume for each container are rather infrequent, compared to the filling times to reach the respective ideal volumes. In addition, the rate at which containers should be emptied varies between containers. There is a class imbalance between the successful emptying actions (and corresponding rewards) and the do-nothing actions making the distributions of actions highly asymmetric, with important actions being relatively rare.
\end{compactitem}

\section{Reinforcement Learning Problem Formulation}\label{sec:RL formulation}

In this section, the container management problem referenced in Section~\ref{sec:Problem Description} is recast within the framework of a Markov Decision Process (MDP), with a corresponding digital twin designed in Python using gymnasium \cite{towers_gymnasium_2023}. This adaptation retains the crux of the actual problem while deliberately omitting edge cases such as facility downtime and the duration of minor processes for simplicity. The state space is carefully crafted, incorporating all critical elements required for effective decision-making by the agent. By grounding the model's parameters in authentic data, the MDP offers a precise representation of the real-world problem at hand.

\subsection{State space.}

The system's state at a timestep $t$ ($s_t$), encompasses several key components. These include container volumes  $\{v_{i, t}\}_{i = 1}^{n}$ bounded by predefined minimum and maximum volumes to maintain operational constraints. $\{p_{j, t}\}_{j = 1}^{2}$ captures the normalized (by timestep) time until each of the two processing units becomes available, ensuring that the system can anticipate and plan for processing availability. $\{b_{k, t}\}_{k = 1}^{n}$ is a binary representation indicating which containers are currently being emptied, providing immediate insight into the immediate container status, rewards from the previous timestep $\{r_{l, t-1}\}_{l = 1}^{n}$, used to integrate feedback from past decisions, and the ideal volumes $\{pv_{m}\}_{m = 1}^{n}$ for each container guiding the agent towards maintaining optimal material levels. The indices $i$, $j$, $k$, $l$, and $m$  denote the container or PU identifiers.  This representation is expressed as follows:

\begin{equation}
    s_t = (\{v_{i, t}\}_{i = 1}^{n}, \{p_{j, t}\}_{j = 1}^{2}, \{b_{k, t}\}_{k = 1}^{n}, \{r_{l, t-1}\}_{l = 1}^{n}, \{pv_{m}\}_{m = 1}^{n})
\end{equation}
This formulation ensures an adequate reflection of the system's dynamics, incorporating both the containers' status and the operational state of PUs, thereby facilitating the decision-making process.

\subsection{Action space.}
At any given time $t$, the agent has two primary choices: (i) to refrain from emptying any container, effectively taking no action and letting the volume increase, or (ii) to empty a specific container and process its contents. The \textit{do-nothing} action is denoted by $0$, while the action of emptying a container and processing its content is denoted by the index $i$ of the container. Thus, the action $a_t$ falls within the set $\{0, 1, \dots, n\}$, where $n$ represents the total number of containers. Depending on the availability of the PUs the action taken is either successful or unsuccessful for emptying and is reflected through the volumes, status of PUs and time features in the state, and the reward received. 

\subsection{Environment Dynamics}
In this section, the dynamics of the volume of material in the containers, the PU model, as well as the state update are discussed.

\subsubsection{Container filling rates.}

In addressing the variability of container fill rates due to the inconsistent nature of plastic material inflow, which ranges from solid lumps to irregular flows, leading to uneven accumulations within containers, a stochastic model is employed. This model accounts for the erratic yet on average linear increase in volume over time through a random walk mechanism with drift. Specifically, the volume update for container $i$ at time $t + 1$ is modeled as:
\begin{equation}
    v_{i, t+1} = \max(0, \alpha_i + v_{i, t} + \epsilon_{i, t}),
\end{equation}
where $\alpha_i$ represents the average volume increase rate, and $\epsilon_{i, t}$ is a normally distributed random variable representing measurement noise. This approach captures real-world fluctuations in fill rates while simplifying the model to facilitate analysis and simulation. Upon the action of emptying a container, its volume is instantaneously reduced to zero in the subsequent timestep. This simplification contrasts with the gradual volume reduction observed in actual scenarios but aligns with our dataset's indication that the emptying process completes within the duration of a single timestep, set at 60 seconds in this study. This modelling choice effectively bridges the gap between the discrete-time model used for simulation and the continuous nature of the emptying process observed in operational environments.

\subsubsection{Processing unit dynamics.}

The transformation time by a Processing Unit (PU) for material volume $v$ in a container depends linearly on the producible products from $v$, expressed as $\lfloor v / b_{ij} \rfloor$. This is detailed by the function $g_{ij}$ in Equation~\eqref{eq:pu_model}, incorporating $b_{ij}$ for product size, $\beta_{ij}$ as the PU's activation time, and $\lambda_{ij}$ for time per product. Here, $i$ refers to the container index, and $j$ to the specific PU index, indicating that the parameters are specific to each container-PU combination. If PUs are occupied, containers awaiting processing continue to accumulate material, highlighting the system's operational constraints and the need for efficient PU management.

\begin{equation}\label{eq:pu_model}
g_{ij}(v) = \beta_{ij} + \lambda_{ij} \lfloor v / b_{ij} \rfloor \enspace .
\end{equation}

\section{Reward Tuning}\label{sec:reward_tuning}

In addressing the complexities of dynamic decision-making environments, the formulation of reward functions is a critical component. This section presents the case for Gaussian-based reward mechanisms, chosen for their smoothness and ability to accommodate the infrequent nature of container-emptying actions and the uncertainties within state space parameters. Gaussian rewards, characterized by their resilience to noise, ease of interpretation, and smooth gradient provision, are selected to mitigate the effects of measurement inaccuracies and focus on the aggregate effects of the underlying phenomenon. The strategies are presented in three distinct subsections: Simple Gaussian Reward, Custom Reward, and Precision Reward.

\subsection{Simple Gaussian reward}

The reward $r(s_t, a_t, PU_{status})$ is calculated based on the action taken, the current volume of a container, the volume at the next time step,  and whether a PU is free to empty the container implicitly contained in $s_t$. A Gaussian distribution centred around the ideal volume for emptying a given container with parameters defining the peak volume (\(p\)), peak height (\(a\)), and peak width (\(w\)) was used. 
The complete reward function $r(s_t, a_t, PU_{status})$ is summarized in Algorithm~\ref{alg:simple gaussian reward}. It fosters emptying containers at or close to their optimal volumes, where the width (standard deviation) $w$ controls the required precision encoded by the reward. A wider peak enables fast initial learning, while a narrow peak focuses on fine-tuning.

\begin{algorithm}
\caption{Gaussian Reward Function for Emptying Decision}
\label{alg:simple gaussian reward}

\SetKwInOut{Input}{input}
\SetKwInOut{Output}{output}

\Input{Action $a_t$, Current volume $v_{t}$, PU availability $PU\_status$, penalty $\rpen$, ideal volume $v_{i}$, Height of the peak $h$, Width of the peak $w$ 
}
\Output{Reward $r_t$ between $\rpen$ and 1}

\BlankLine
\If{$a_t > 0$}{
  \If{$v_{t} = 0$ \textbf{or} not $PU\_status$}{  
     $r_t \leftarrow \rpen$ 
  }
  \Else{
     $r_t \leftarrow (h - \rpen) \cdot \exp\left(- \frac{(v_{t} - v_{i})^2}{2 w^2}\right) + \rpen$ 
  }
}
\end{algorithm}

In principle, this reward encodes everything the agent needs to know about the problem at hand.

\subsection{Custom Reward}{\label{custom_reward}}

Here the custom reward function built on the Gaussian reward is introduced to navigate the complexities of dynamic resource allocation and operational optimisation. This reward is a sum of action rewards plus bonuses, positional rewards, and episode termination rewards to enhance the learning efficiency of the agent. 

\textbf{Action Reward}, which is nothing but the Simple Gaussian Reward Function, described earlier is used to quantify the efficacy of an action at a timestep based on the container's proximity to an ideal volume. It applies only to (supposedly rare) emptying actions. The conditional nature of this reward ensures that meaningful actions-—those contributing to the actual emptying behaviour of the agent--are incentivised, while actions that do not align with constraints are penalised.  

\textbf{Positional Reward} is used to encourage the system towards an ideal operational state across all containers, not just the one being acted upon. It's designed to address the challenge of slow filling rates across the containers, which leads to sparse action rewards within the operational timeframe. For example, there are containers for which it takes up to 300 timesteps of 60 seconds (or five hours) to reach the ideal emptying volume. These rewards need to be propagated back through a long history of zero-actions that contributed to emptying at the right moment. By assigning a reward to each container based on its volume relative to an ideal operational state, and importantly, doing so irrespective of the agent's actions \emph{at every time step}, the reward signal ensures a constant nudge towards optimal efficiency across all containers.

The positional reward for a given container not acted upon at time step \(t\), \(r_{\text{positional},j}\), is calculated as follows:
\begin{equation}
r_{\text{positional},j} = \begin{cases} 
1 - \left| \frac{( v_{i} - v_{j,t})}{ v_{i}} \right|^{0.5}, & \text{if } v_{j,t} \leq  v_{i}, \\
-0.1, & \text{otherwise}.
\end{cases}
\end{equation}


The cumulative positional reward for all containers is calculated as

\begin{equation}
r_{\text{positional}} = \sum_{j=1}^{n} r_{\text{positional},j} 
\end{equation}
where \(n\) is the total number of containers.

\textbf{Episode Termination Reward} is a critical component designed to ensure the agent learns to avoid prematurely ending an episode by reaching an unsafe state. It aims at discouraging the agent from allowing any container to exceed its physical limit of 40 volume units, a crucial safety consideration.

The episode termination reward, \(r_{\text{termination}}\), is defined as follows:
\begin{equation}
r_{\text{termination}} = \begin{cases} 
0.2, & \text{small positive reward for not ending the episode prematurely}, \\
-30, & \text{if any container's volume exceeds the safety limit of 40}.
\end{cases}
\end{equation}

\subsection{Precision Reward}

While the Gaussian reward in principle encodes the full goal of emptying at the optimal volume, its smooth and rather flat top can make fine-tuning difficult. Therefore, we introduce an additional incentive for the agent to empty at the ideal volume by rewarding within a narrowly defined optimal range and penalizing deviations:

\begin{algorithm}
\caption{New Reward Style for Encouraging Ideal Volume}
\label{alg:new_reward_style}

\SetKwInOut{Input}{input}
\SetKwInOut{Output}{output}

\Input{Current volume $v_{t}$ of a container, Ideal volume $v_{i}$}
\Output{Reward $r_t$}

\BlankLine
\eIf{$-0.5 +  v_{i} < v_{t} < 0.5 +  v_{i}$}{
    $r_t \leftarrow 1.0$ 
}{
    $r_t \leftarrow -0.1$ 
}

\end{algorithm}

\begin{figure}
  \centering
  \begin{minipage}{0.31\textwidth}
    \includegraphics[clip, trim=0.1cm 0.1cm 0.1cm 0.1cm, width=\textwidth]{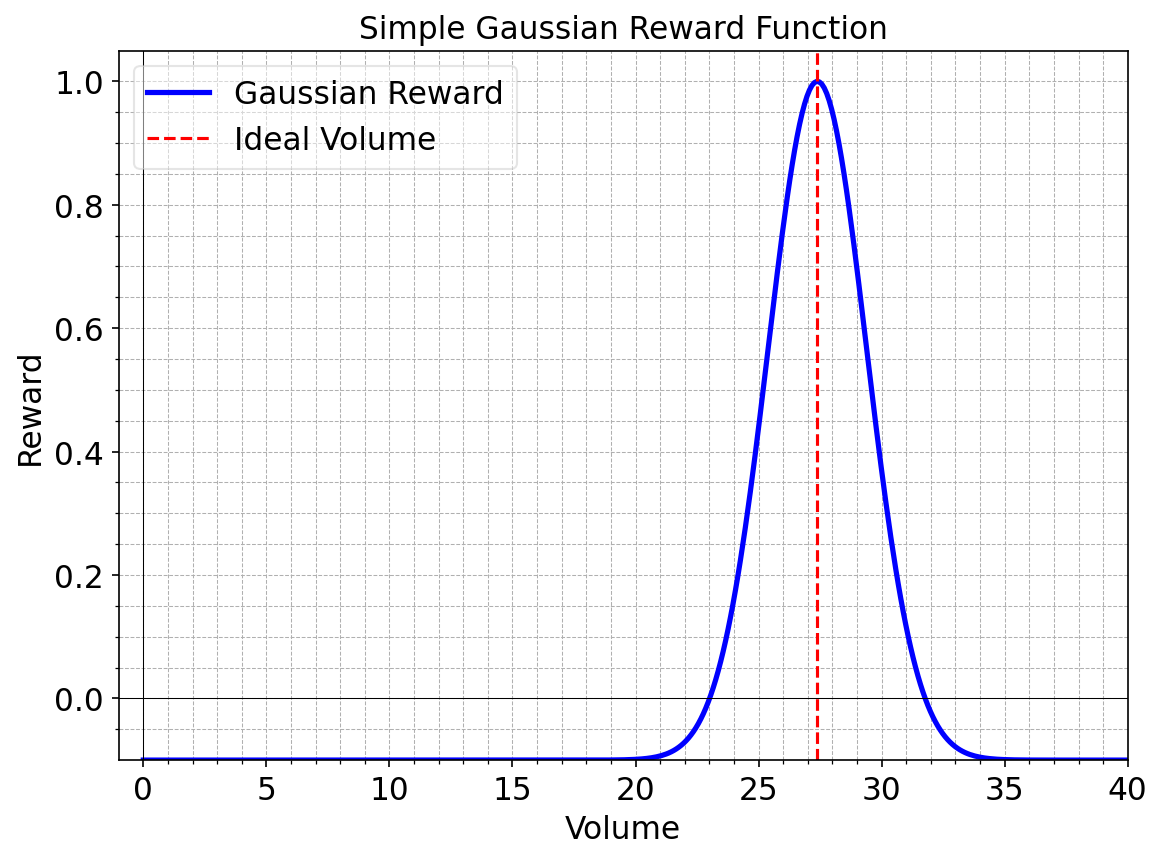}
  \end{minipage}
  \hfill 
  \begin{minipage}{0.31\textwidth}
    \includegraphics[clip, trim=0.1cm 0.1cm 0.1cm 0.1cm, width=\textwidth]{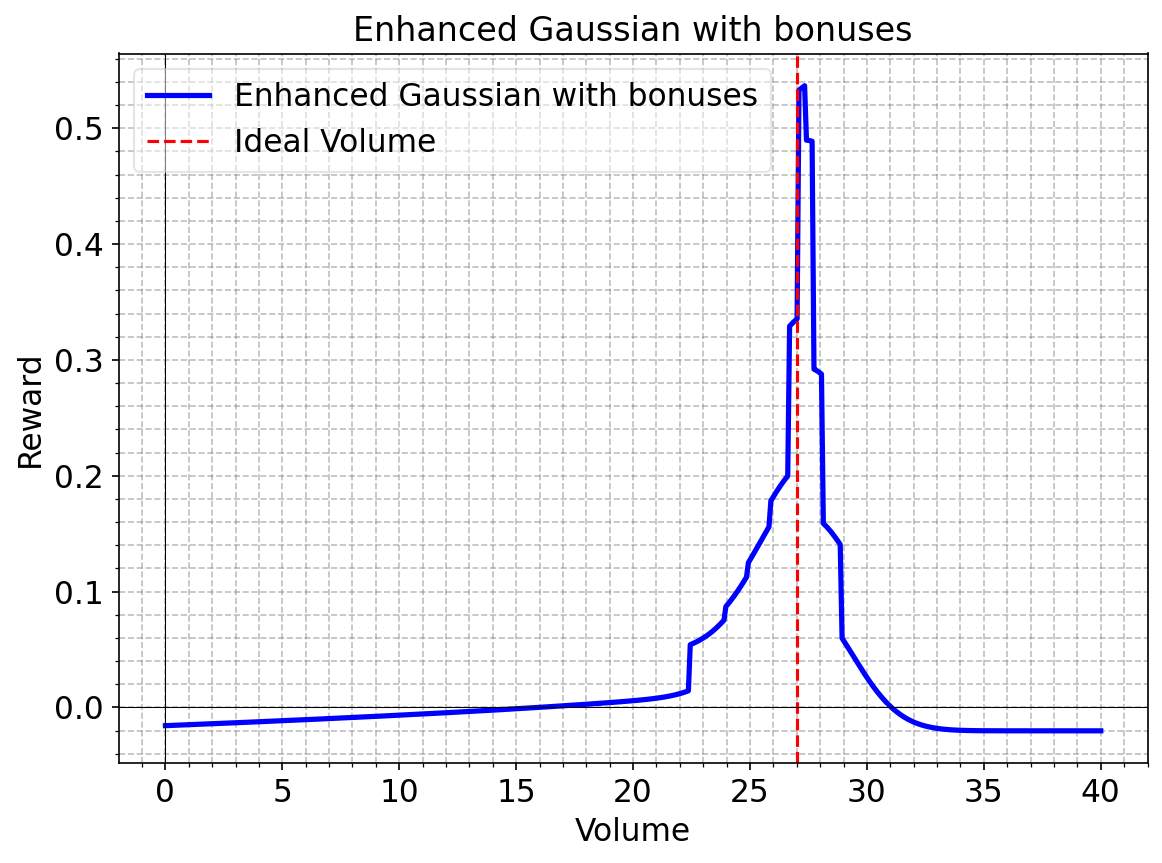}
  \end{minipage}
  \hfill 
  \begin{minipage}{0.31\textwidth}
    \includegraphics[clip, trim=0.1cm 0.1cm 0.1cm 0.1cm, width=\textwidth]{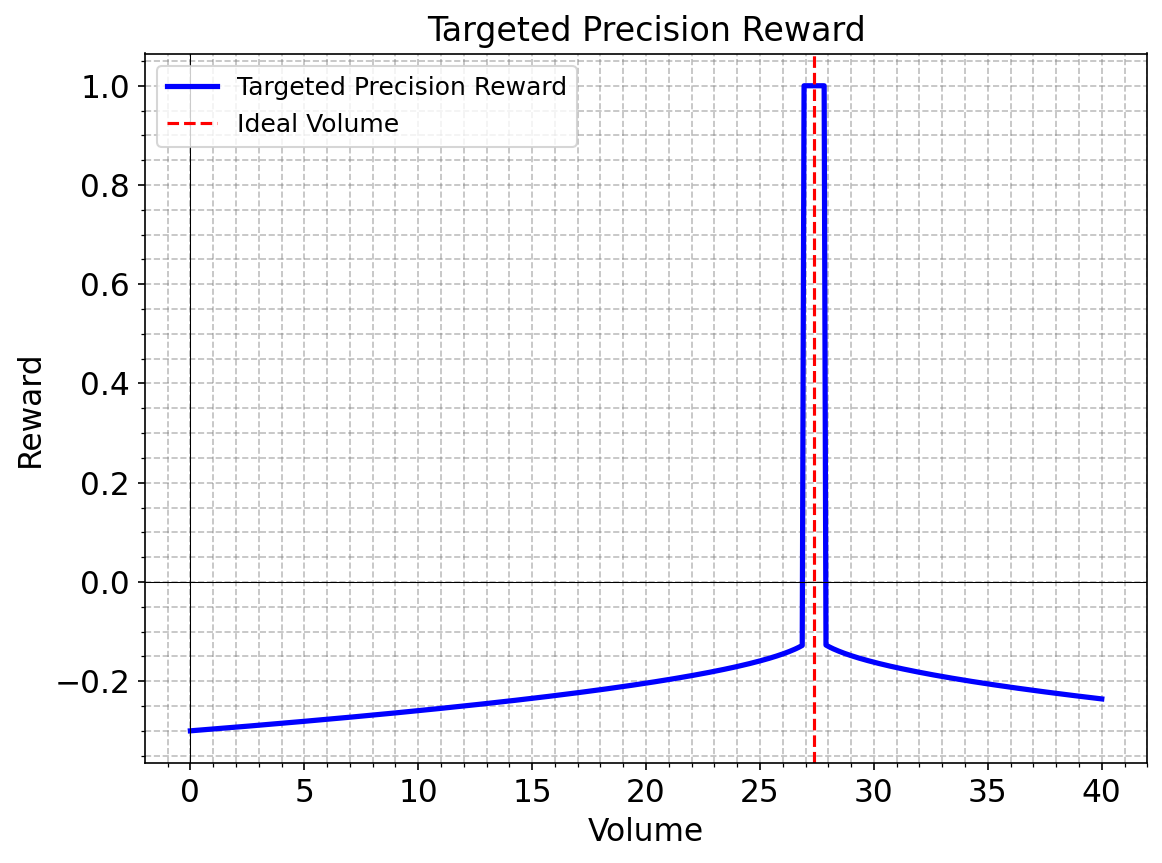}
  \end{minipage}
  \caption{Plots showing the various reward functions: Simple Gaussian (left), Custom Reward (centre), and Precise Reward (right).}
  \label{fig:reward_functions_comparison}
\end{figure}

\section{Methodology}\label{sec:methodolody}

In the section, a scaffold approach to developing reinforcement learning agents is presented. This methodology, encompassing five phases, progressively introduces increased complexity, systematically enhancing agent competency. By employing a phased curriculum learning strategy, the agent's evolution is carefully curated, ensuring a gradual ascension to operational adeptness. In the following, we go through each phase, starting from its goals, and explaining the measures for reaching these goals.

\paragraph{\textbf{Phase 1: Foundational Training.}} The goal of this phase is to learn a first non-trivial policy for a simplified version of the task. In particular, the agent must learn to perform the zero-action most of the time, while executing rare but critical emptying actions when a container volume comes close to the optimal volume. In this initial phase, the random walk for filling containers is disabled, resulting in a deterministic and easily predictable environment. Furthermore, there are as many PUs as containers. That's a very unrealistic setting that in effect removes all dependencies between containers reaching peak volume at the same time. Here, the Custom Reward is implemented. Actions are taken every 30 seconds, and episodes are as short as 25 time steps. The initial states are designed such that there are at least 8-11 emptying actions that lead to positive rewards and the rest being do-nothing actions. This helps the initial agent in dealing with the class imbalance problem alluded to in section~\ref{sec:Problem Description}.

\paragraph{\textbf {Phase 2: Refinement.}} In this stage, the Precision Reward is added for a budget of 1 million timesteps. All other settings are retained from the first phase. The introduction of Precision Reward aims to refine the agent's accuracy in actions, within a deterministic setting, enhancing the precision in decision-making.

\paragraph{\textbf{Phase 3: Penalizing Greedy Actions.}}
This training stage aims at altering the policy so that it becomes more energy efficient. This is achieved by slightly penalizing all emptying actions. The reasoning is that PUs are a scarce resource, and emptying a container twice instead of once blocks the PU for an unnecessarily long time, and it also uses more energy than processing more of the same material in one go. This adjustment is intended to augment the agent's greedy nature to accumulate more episodic rewards by prematurely emptying a container too many times as opposed to optimizing for the least number of emptying actions per episode along with precise emptying actions.

\paragraph{\textbf{Phase 4: Inject Real-world Complexity.}}
Now that the basic behaviour is as intended, the agent is ready to be exposed to the true complexity of the environment. The budget for this phase is 0.5 million timesteps. Timesteps are extended to 60 seconds, and the episodes are up to 600 timesteps long. In addition, the stochastic filling rates and resource constraints are introduced. This results in significant changes in the environment dynamics (noisy dynamics, failing emptying actions if all PUs are busy), which must be accounted for with corresponding changes in the value function estimator. To achieve this, the policy network is frozen and only the value network is trained as both of these are distinct neural nets not sharing network parameters \cite{Wang2020-qc}. This phase is tailored to tune the agent’s value network to the unpredictable and constrained operational dynamics.

\paragraph{\textbf{Phase 5: Fine-tuning with Reduced KL Constraint.}} This phase concludes the curriculum, mirroring Phase 4’s operational parameters but reinstating the Precision Reward. Gradually unfreezing the policy with a low KL constraint (limiting the Kullback Leibler divergence between action distributions) allows controlled exploration around the learned policy within the complex environment introduced in Phase 4.

To address operational constraints and edge cases effectively, action masking was incorporated after the curriculum learning phases for inference on the test environment. This adaptation ensures alignment with the plant's structural and operational limitations, where the disposition of containers relative to Processing Units (PUs) is dictated by the proximity to conveyor belts. Given this configuration, only specific containers can be targeted for emptying based on their accessibility to certain PUs. Additionally, dynamic reduction of the action space is facilitated through action masking \cite{Huang2020-eg}, particularly when all PUs are engaged, enhancing decision-making efficiency.

\begin{table}[ht]
    \caption{Summary of curriculum learning phases and their parameters}
    \centering
    \resizebox{0.99\textwidth}{!}{
    \begin{tabular}{|c|p{1.2cm}|c|c|p{2.5cm}|c|c|}
        \hline
        Phases & Budget & TimeSteps (Ts) & Ep-len (in Ts) & Reward & Fill.\ rates & Resource Constraint (PUs) \\
        \hline
        Phase-1 & 1.5 & 30 & 25 & Custom  & Deterministic & No \\
        Phase-2 & 1 & 30 & 25 & Precision  & Deterministic & No \\
        Phase-3 & 1 & 30 & 25 & Precision with penalty for positive actions & Deterministic & No \\
        Phase-4 & 0.5 & 60 & 600 & Policy frozen and Precision  & Stochastic & Yes \\
        Phase-5 & 0.5 & 60 & 600 & Precision Reward & Stochastic & Yes \\
        \hline
    \end{tabular}
    }
    \label{tab:curriculum_learning_phases}
\end{table}

\section{Experimental Evaluation}\label{sec:results}

The goal of this section is to empirically verify the effectiveness of the proposed approach. We aim to achieve this by answering two research questions.  

\paragraph{Research question 1:} Is the curriculum approach effective? To this end, we compare the PPO-Curriculum learning (PPO-CL) agent trained in five phases for all three criteria agents with our first baseline. This is a naive  PPO agent (PPO-volume criteria), optimized only for one of the three criteria, i.e., optimal emptying volumes encoded by training with a Simple Gaussian reward. 

\paragraph{Research question 2:} Does the proposed approach offer a real-world benefit? For this, we compare PPO-CL with our second baseline, which is a meticulously hand-crafted analytical agent labelled as a Optimal Analytic agent. This agent is close to the semi-automated decision policy currently deployed at the waste sorting facility. It is optimal in the sense that it empties containers at their precise target volumes if possible, and that it performs emergency emptying at a volume of 37 (out of 40) to avoid overflow. However, it does not take interactions of containers competing for PUs into account.

With respect to metrics, cumulative reward values are not necessarily the most relevant performance measure. Therefore we investigate different aspects like safety, energy-saving behavior and timing precision separately, in order to arrive at conclusions that are meaningful not only from an RL perspective but also from an application point of view.

In the spirit of open and reproducible research, we make our source code available via an anonymous repository.%
\footnote{\url{https://gitlab.com/anonymousppocl1/ppo_paper.git}}
The repository contains a script for reproducing all results presented in this section.

\subsection{Experimental Setup}

The PPO-CL (five phases) and PPO-volume criteria agents undergo training within environments with eleven containers and two PUs. These are characterized by a 60-second time-step and a 600-episode length, with default settings maintained for other hyperparameters. To evaluate stability, fifteen independent training runs are conducted with distinct random seeds for each agent. Our second baseline, the Optimal Analytic agent operates with a fixed logic. Hence it is designed, not trained. Its main criterion for emptying decisions is the evaluation of the proximity of each container's volume to its ideal state.  It also prioritizes actions based on operational conditions such as PU availability and the plant's physical layout (e.g., container alignment with conveyor belts). To ensure a fair comparison, we test the PPO-CL agent and both baselines on the same test environment.

\subsection{Results}

\begin{figure}
  \centering
  \begin{minipage}{0.49\textwidth}
    \includegraphics[clip, trim=2.3cm 2.3cm 0.3cm 0.3cm, width=\textwidth]{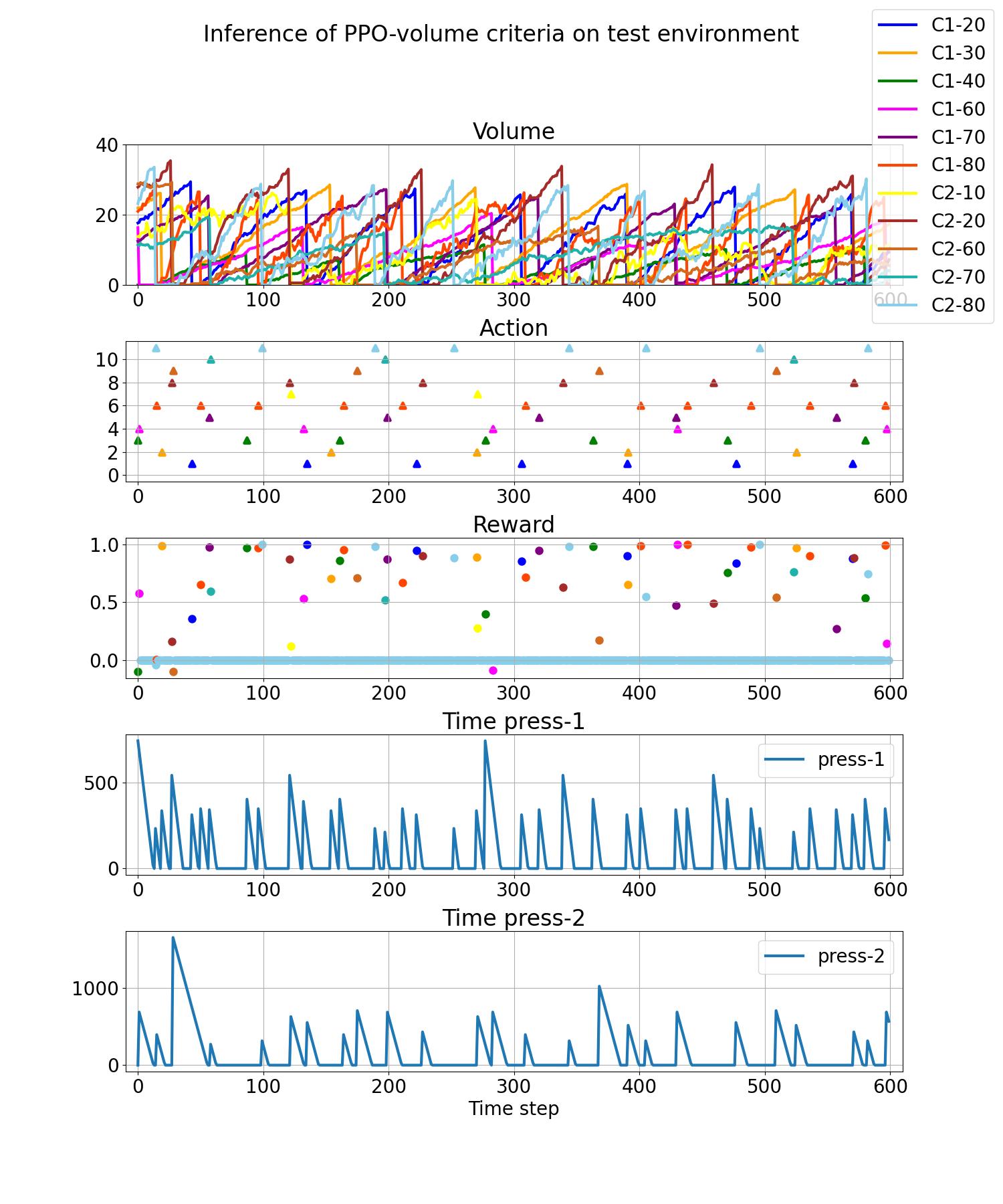}
  \end{minipage}
  \hfill 
  \begin{minipage}{0.49\textwidth}
    \includegraphics[clip,trim=2.3cm 2.3cm 0.3cm 0.3cm, width=\textwidth]{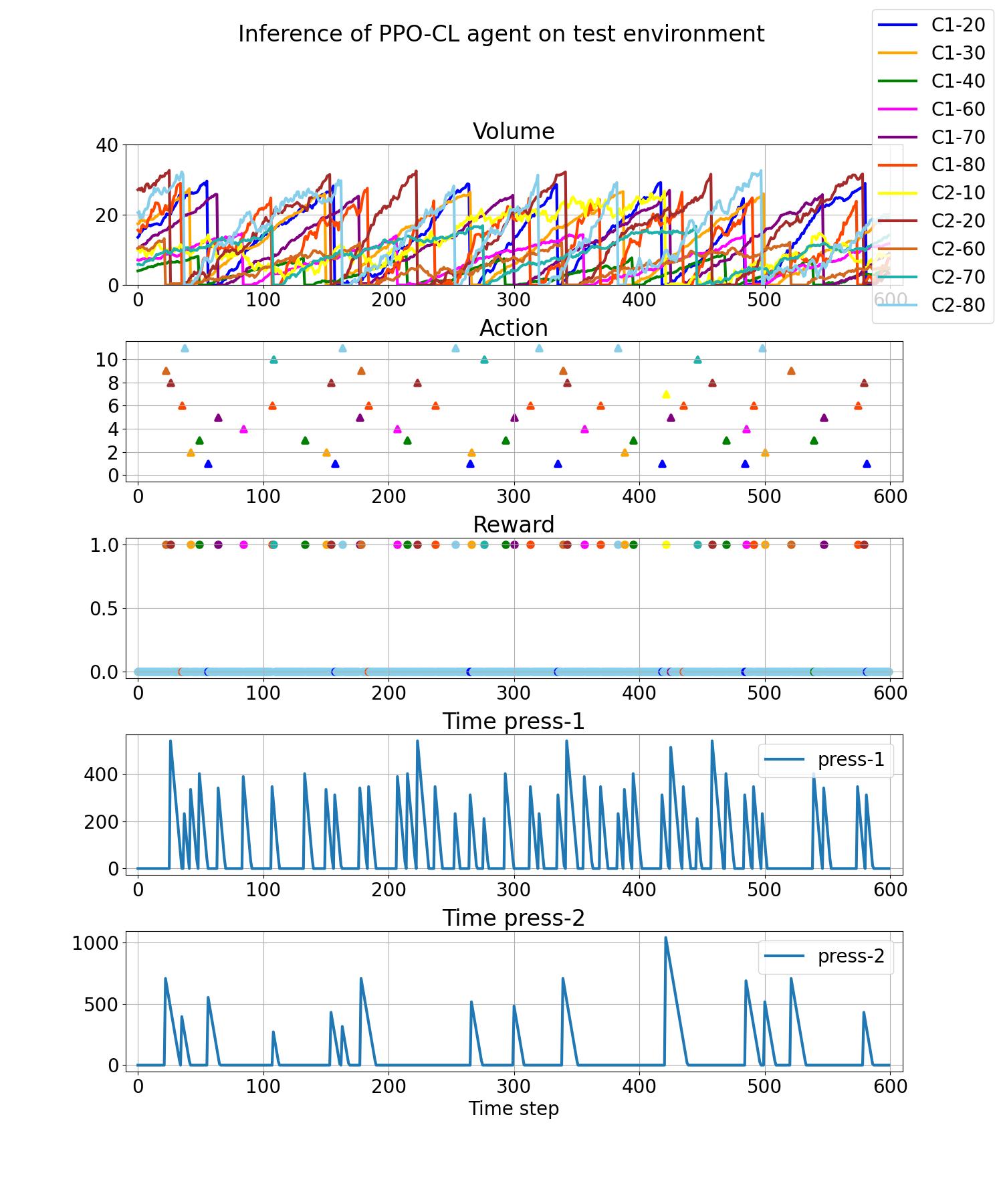}
  \end{minipage}
  \caption{A single rollout (best agent out of 15) of the PPO-CL (left) and PPO-volume criteria on a test environment with 11 containers. Displayed are the volumes, emptying actions, rewards, and time to process by PU-1 and PU-2.
    \label{fig:inference}
  }
\end{figure}

In this section, we present the empirical results. Figure \ref{fig:inference} shows a single rollout with the best policy produced by both PPO agents. Our visualization tracks container volumes, actions, and PU utilization over time, offering a more nuanced analysis than reward values alone. The volume chart shows different containers being emptied at around their respective ideal volumes and the frequency of peaks in the time chart for PUs gives a sense of how they compete for the resources. The key insight here is that although the PPO-volume criteria agent seems to show reasonable behaviour, its emptying decisions are not properly timed, as is evident from the large reward fluctuations. In contrast, the PPO-CL agent consistently achieves close-to-optimal rewards. 

\subsubsection{Empyting Decision Quality}

\begin{figure}
  \centering
  \begin{minipage}{0.70\textwidth}
    \includegraphics[clip, trim=0.3cm 1.6cm 0.3cm 0.4cm,  height=0.33\textheight]{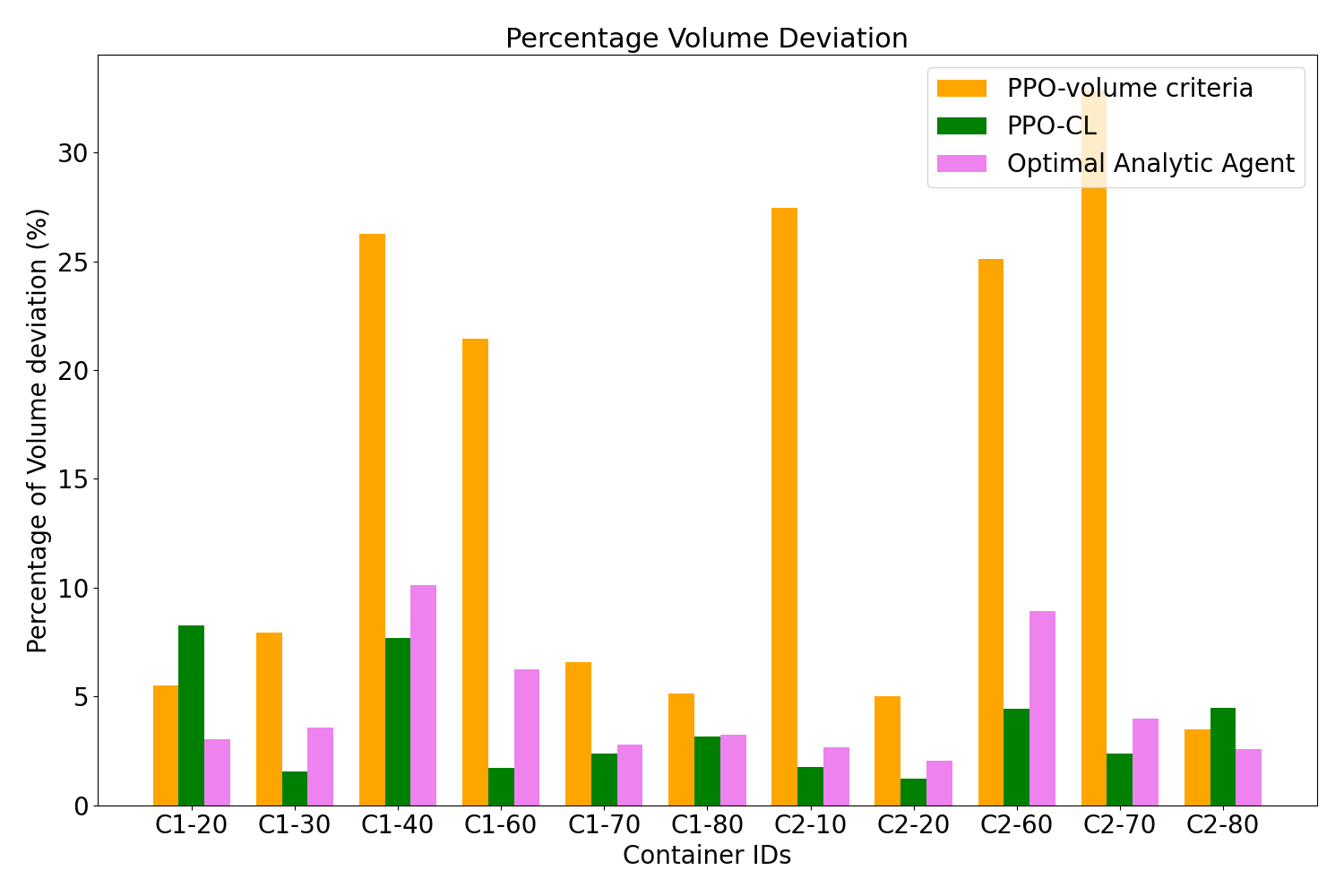}
  \end{minipage}
  \caption{ Comparision of percentage volume deviation across all containers, collected over 15 rollouts of the best policy for both PPO agents
    \label{fig: percentage volume deviation}
  }
\end{figure}

Here we compare all three agents for the quality of emptying decisions. For the trained PPO agents, these metrics represent the best out of 15 independent training runs. Table \ref{tab:best_seeds_metrics} shows the average inference volume deviation (deviation between actual and ideal container volume) across all containers. Figure \ref{fig: percentage volume deviation} compares the container-wise percentage volume deviation of all agents.  

The first baseline, the PPO-volume criteria agent, performs poorly with the highest deviation. The PPO-CL agent and the analytical agent achieve much smaller deviations, while the former does better than the latter. This is also evident from looking at their performance individually for each container. The PPO-volume criteria agent, although trained only for optimizing volume, has higher than 20\% volume deviation for five containers, which is quite excessive. In contrast, the PPO-CL agent has the lowest deviation values for the majority of the containers.

\subsubsection{Safety and Energy savings}

Table \ref{tab:best_seeds_metrics} also shows the total number of emptying actions. Both baseline agents take the same number of emptying actions, while PPO-CL gets along with fewer actions, consequently emptying containers at higher volumes. This behavior results in better resource utilization, in terms of PU occupation time, and also in terms of energy.

\begin{figure}
  \centering
  \begin{minipage}{0.49\textwidth}
    \includegraphics[clip, trim=0.3cm 1.0cm 0.3cm 0.3cm, width=\textwidth]{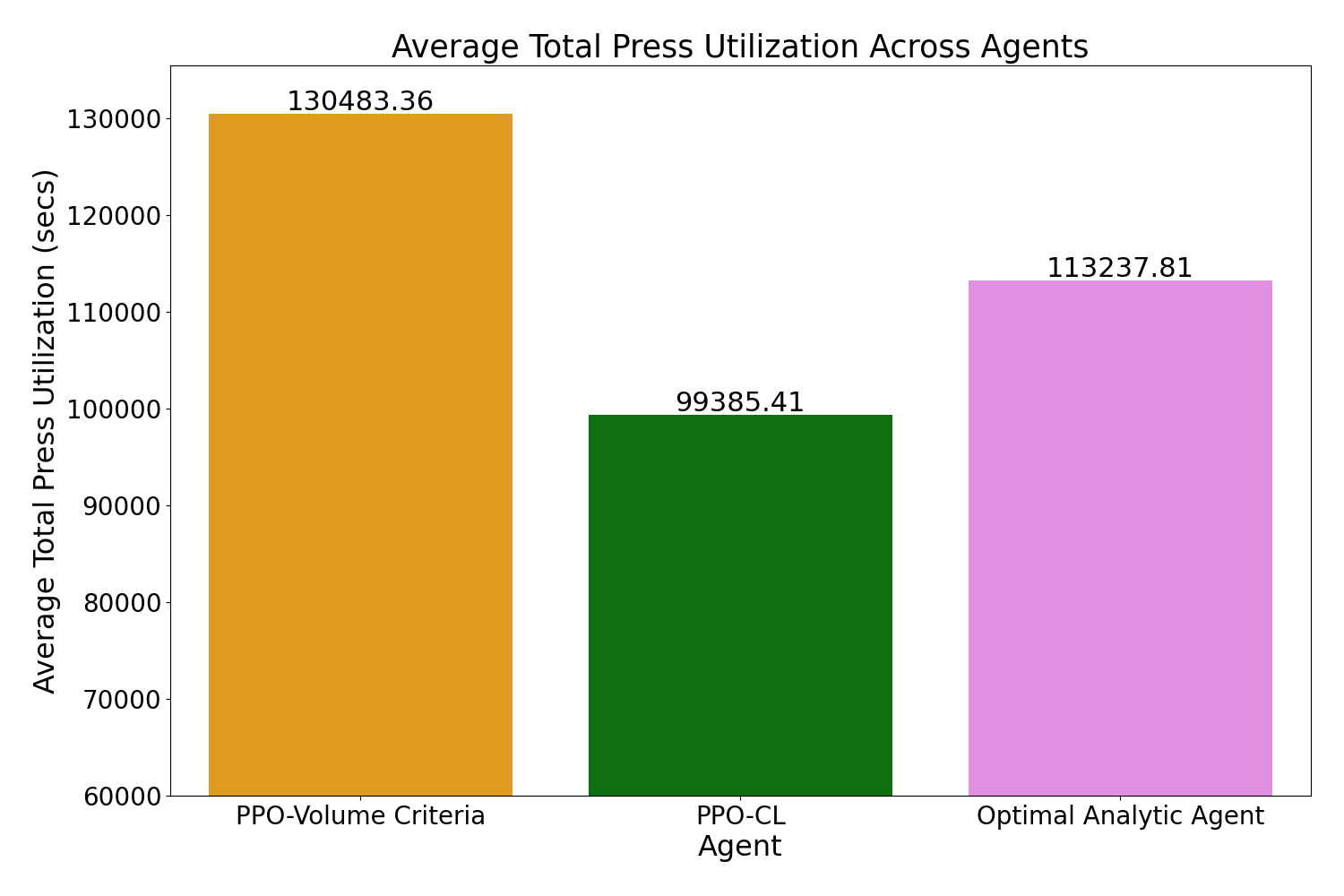}
  \end{minipage}
  \hfill 
  \begin{minipage}{0.49\textwidth}
    \includegraphics[clip, trim=0.3cm 1.0cm 0.3cm 0.3cm, width=\textwidth]{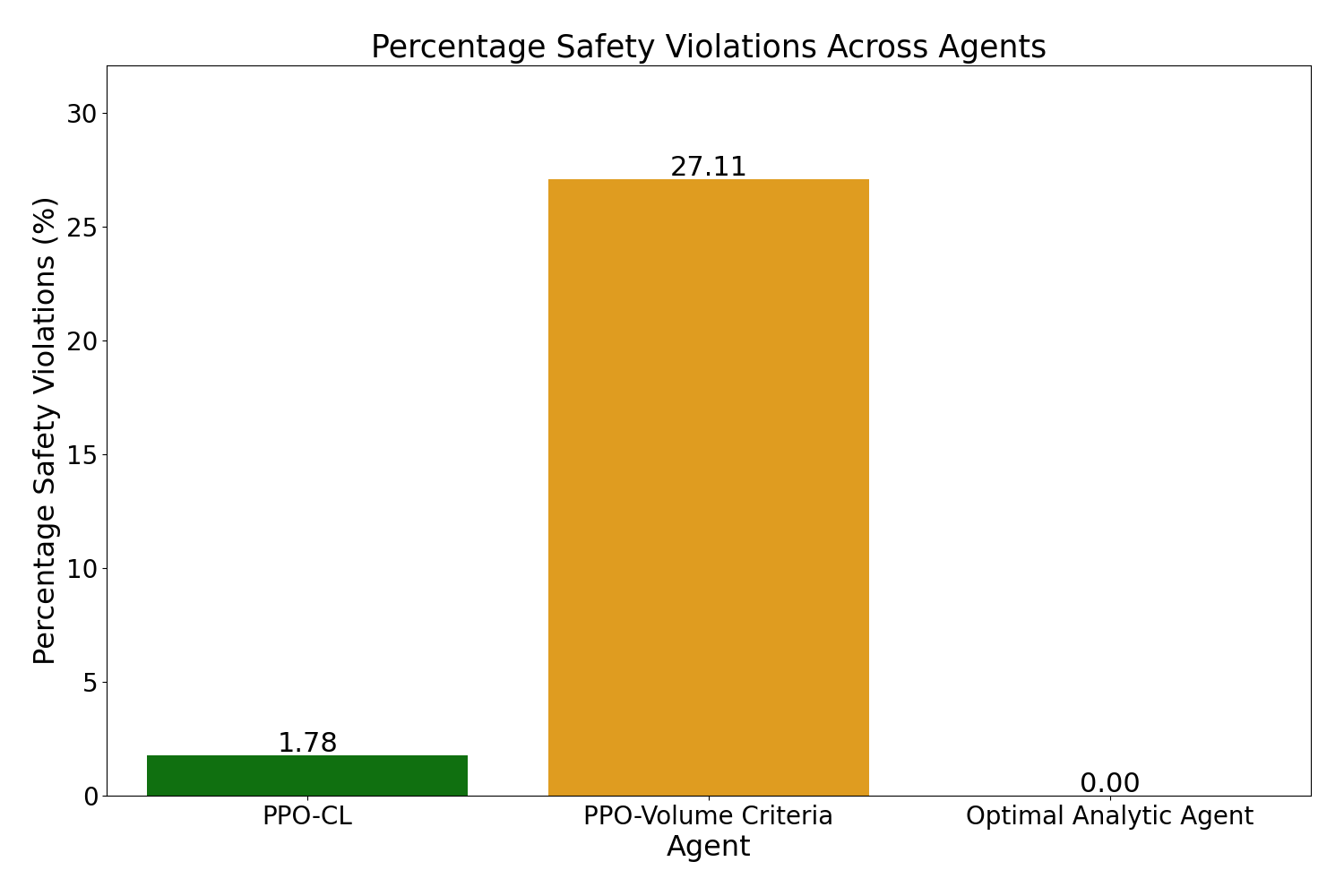}
  \end{minipage}
  \caption{Comparison of key performance metrics across different agents, collected over $15$ rollouts of the best policy for both PPO agents. The left figure presents the average total PU utilization across agents. The right figure details the percentage of safety violations
    \label{fig:performance_safety_comparison}
  }
\end{figure}

Figure \ref{fig:performance_safety_comparison} further highlights the PPO-CL agent's strengths in energy efficiency and safety compliance.  On the left, we see significant energy conservation: the PPO-CL agent utilizes the PU 12\% less than the Optimal Analytic Agent and 24\% less than the PPO-volume criteria agent. The right graph underscores the PPO-volume criteria agent's alarming safety violations: 27.11\% of its actions exceed the critical volume limit of 40. This highlights the risks of training an RL agent from scratch in complex environments with safety constraints. In contrast, the Optimal Analytic agent exhibits zero violations due to its hard-coded limit of 37, acting as a safety benchmark but lacking adaptability.  The PPO-CL agent, however, achieves a remarkable 1.7\% violation rate. 

Overall, the PPO-CL agent demonstrates a superior balance. The results suggest that phased learning strategies offer distinct advantages in complex, multi-criteria decision-making environments, particularly in industrial settings where precision, adaptability, and efficiency are crucial.

\begin{table}[ht]
\caption{Performance metrics for both RL agents (best seed out of 15) and analytic agent for 15 rollouts on the same test environment: emptying actions and average inference volume deviation}
  \centering
  \resizebox{0.75\textwidth}{!}{
  \begin{tabular}{|c|c|c|}
    \hline
    Agent & Emptying Actions & Avg. Inf Vol Deviation ($\pm$ Std Dev) \\ 
    \hline
    PPO-volume criteria & $62$ & $15.16 \pm 10.80$ \\ 
    Optimal Analytic Agent & $62$ & $4.47 \pm 2.61$ \\ 
    PPO-CL & $56$ & $3.55 \pm 2.33$ \\ 
    \hline
  \end{tabular}
  }
  \label{tab:best_seeds_metrics}
\end{table}

\begin{figure}
\centering
\includegraphics[clip, trim=0.1cm 0.1cm 0.1cm 0.1cm, width=0.24\textwidth]{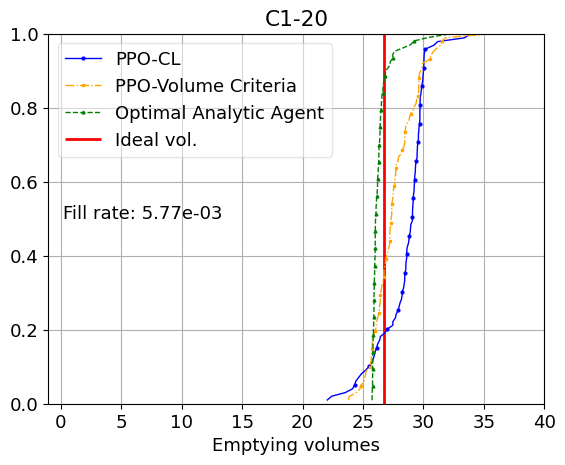}
\includegraphics[clip, trim=0.1cm 0.1cm 0.1cm 0.1cm, width=0.24\textwidth]{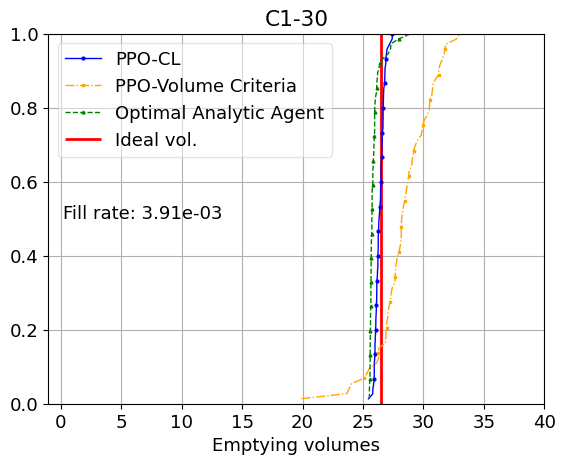}
\includegraphics[clip,  trim=0.1cm 0.1cm 0.1cm 0.1cm, width=0.24\textwidth]{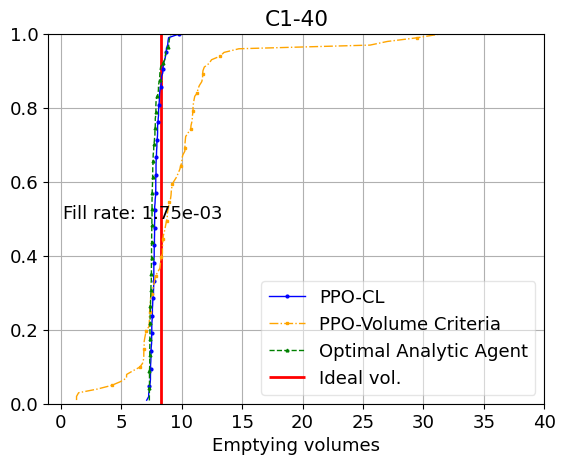}
\includegraphics[clip,  trim=0.1cm 0.1cm 0.1cm 0.1cm, width=0.24\textwidth]{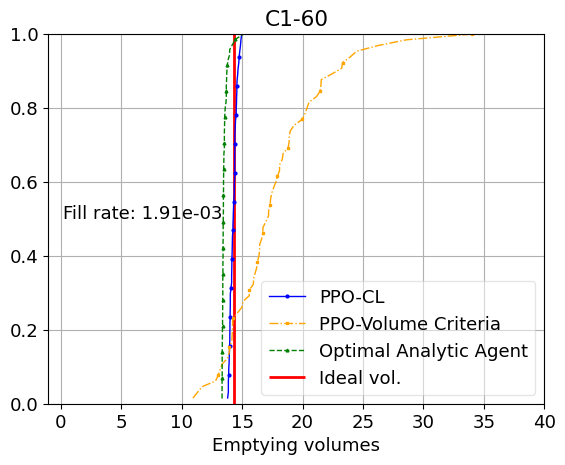} \\
\includegraphics[clip,  trim=0.1cm 0.1cm 0.1cm 0.1cm, width=0.24\textwidth]{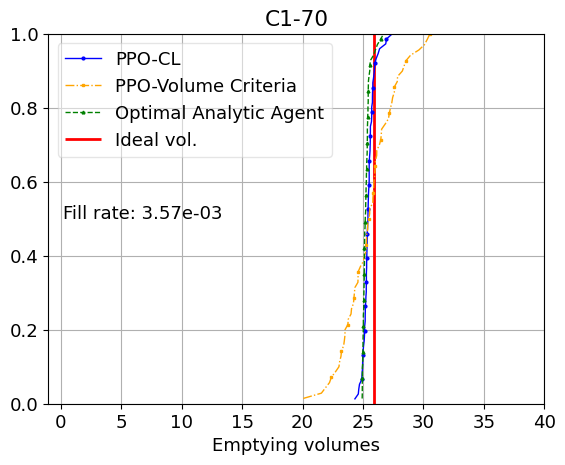}
\includegraphics[clip,  trim=0.1cm 0.1cm 0.1cm 0.1cm, width=0.24\textwidth]{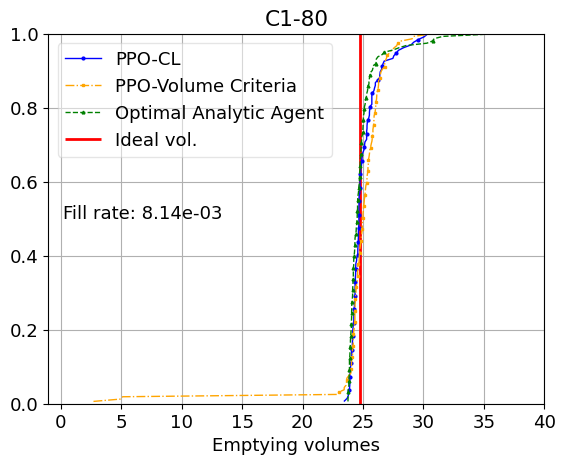}
\includegraphics[clip,  trim=0.1cm 0.1cm 0.1cm 0.1cm, width=0.24\textwidth]{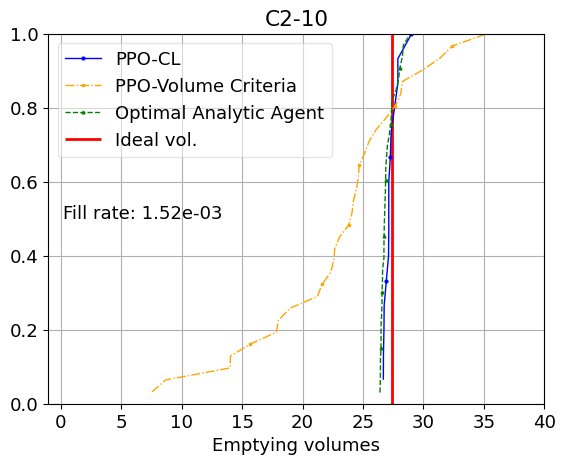} \\
\includegraphics[clip,  trim=0.1cm 0.1cm 0.1cm 0.1cm, width=0.24\textwidth]{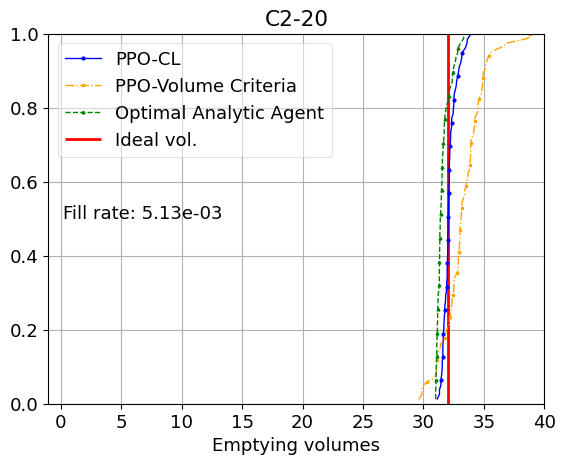}
\includegraphics[clip,  trim=0.1cm 0.1cm 0.1cm 0.1cm, width=0.24\textwidth]{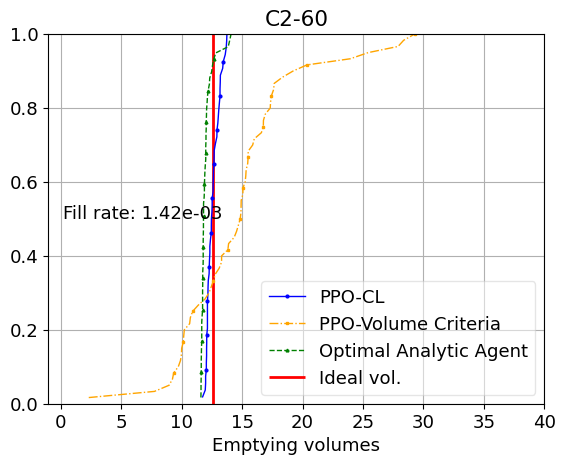}
\includegraphics[clip,  trim=0.1cm 0.1cm 0.1cm 0.1cm, width=0.24\textwidth]{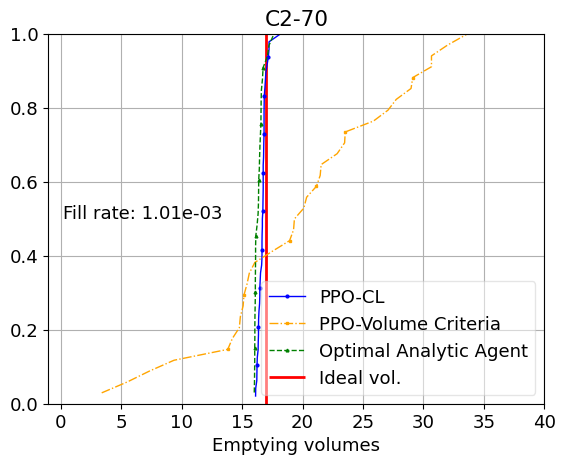}
\includegraphics[clip, trim=0.1cm 0.1cm 0.1cm 0.1cm, width=0.24\textwidth]{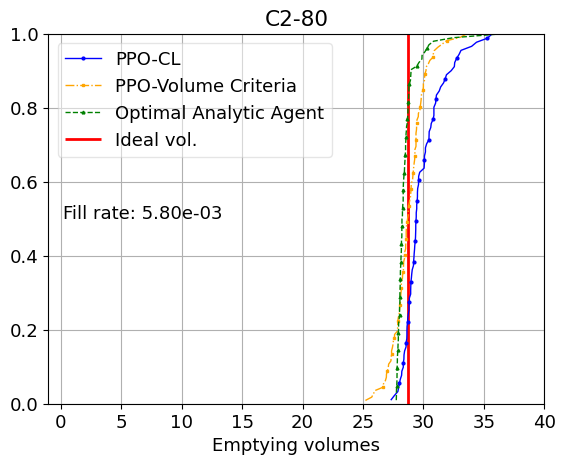}
\caption{ECDFs of emptying volumes of all 11 containers collected over $15$ rollouts of the best policy for PPO-Volume criteria, PPO-CL, and Optimal analytic agent on a test environment. Average fill rates are indicated in volume units per second. 
The derivatives of the curves are the PDFs of emptying volumes. Therefore, a steep incline indicates that the corresponding volume is frequent in the corresponding density.
\label{fig:ecdfs_n5_m2}}
\end{figure}

\subsection{Discussion}

To gain deeper insights into the PPO-CL agent's safety, energy efficiency, and volume management, we further analyze the results from the previous section.

The PPO-CL agent's safety adherence, a critical outcome highlighted by the results, likely stems from two key curriculum design elements: the enforcement of safety constraints  and the early emphasis on safe exploration. Short episode lengths in initial phases (25 timesteps, as seen in Table \ref{tab:curriculum_learning_phases}) encouraged exploration within safe state spaces. Furthermore, the reward structure in Phase 3, which penalizes actions even with positive outcomes, directly incentivizes conservative decision-making, contributing to the agent's reduced PU utilization and energy savings. This flexibility in balancing competing objectives makes our framework particularly suitable for complex industrial settings where safety is of utmost importance. 

To analyze the emptying decision quality of the PPO-CL agent against our two baselines, we examine the empirical cumulative distribution functions (ECDFs) in Figure \ref{fig:ecdfs_n5_m2}.

ECDFs plot the probability, based on observed data from agent actions, that a container is emptied at or below a given volume. Here, steeper curves signify greater consistency in emptying volumes at which a container is emptied across different instances. Our first baseline, the PPO-volume criteria agent, exhibits higher variance in its ECDF, particularly for slower-filling containers (e.g., C1-40, C1-60, C2-10, C2-20). This points to the difficulty of learning effective policies when rewards are delayed due to extended fill times, an issue hindering from-scratch training as discussed in Section \ref{custom_reward}. In contrast, our second baseline, the Optimal Analytic agent, employs a deterministic logic that leads to frequent preemptive emptying. While this ensures no safety violations, it results in inefficient energy usage.

The PPO-CL agent's ECDFs demonstrate remarkable consistency across all containers, regardless of fill rate. This highlights its ability to manage delayed rewards effectively, leading to optimal energy usage with fewer emptying actions per episode – a key advantage emphasized by our results. Crucially, unlike the PPO-volume criteria agent, the PPO-CL agent consistently avoids reaching the critical volume threshold of 40 (evident from all slow-filling containers), demonstrating its strong safety compliance.

\section{Conclusion}\label{sec:conclusion}

In this work, a curriculum learning approach was introduced, designed to progressively enhance the complexity of the environment for training a Proximal Policy Optimization (PPO) agent to navigate the intricacies of real-world industrial operations. Our model not only achieves a harmonious balance among competing operational goals like volume management, energy conservation, and adherence to stringent safety protocols, but it also paves the way for the broader application of such strategies in advanced, adaptive, multi-criteria decision-making environments. 
We believe that with some (industrial) domain expertise available, designing a curriculum is an attractive alternative to using excessive computing for solving hard problems, e.g., by training more complex networks or by applying multi-agent approaches.

We view our achievement as a solid basis for future endeavours. Although our agent performs well in most cases, we have not yet pushed its ability to avoid unsafe behaviour to the maximum. This amounts to avoiding \emph{future} collisions of PU usage requests, which may occur if too many containers reach their ideal volume at the same time. That situation should be counteracted by emptying some containers earlier than usual. The problem is pressing since PUs are expensive units, so facilities are often designed with the smallest possible number of PUs. Designing a systematic solution to this problem will be subject to future research. It may involve forms of (stochastic) real-time planning, as well as further curriculum steps specifically targeted at the collision problem.

For practitioners in the realm of real-world reinforcement learning, we offer two take-home messages. First, it has proven invaluable to evaluate agent performance beyond mere cumulative rewards, using ECDF plots, diverse metrics, and statistical tools. Second, particularly in environments of high complexity, adopting a curriculum-based training approach can help with the learning of meaningful policies, surmounting the limitations faced by vanilla agents.

\subsubsection{Acknowledgements: }

This work was funded by the German federal ministry of economic affairs and climate action through the ``ecoKI'' grant.

\bibliographystyle{splncs04}
\bibliography{bibliography}

\end{document}